\documentclass{article}

\usepackage{times}
\usepackage{graphicx} 
\usepackage{subcaption}

\usepackage{natbib}

\usepackage{algpseudocode}

\usepackage{hyperref}

\usepackage[accepted]{icml2017} 
\usepackage{amsthm}
\usepackage{amsmath}
\usepackage{amssymb}
\usepackage{color}
\usepackage{xspace}
\usepackage{enumitem}
\usepackage{pbox}
\usepackage{array,multirow,graphicx}
\usepackage{dsfont}

\newcommand{\doneasintern}{\ifdefined\isaccepted\textsuperscript{*}Work done during an internship at Google Brain.\fi}

\renewcommand{\comment}[1]{}

\newcommand{\argmax}[1]{\underset{#1}{\operatorname{argmax}}}

\def\deriv{\mathrm{d}}

\renewcommand{\vec}[1]{\boldsymbol{\mathbf{#1}}}

\def\btheta{\vec{\theta}}

\def\bx{\vec{x}}
\def\by{\vec{y}}
\def\yi{y_{i}}
\def\yis{y_{i}^*}

\def\calY{\mathcal{Y}}
\def\calX{\mathcal{X}}
\def\Real{\mathbb{R}}
\def\calP{\mathcal{P}}

\def\byh{\widehat{{\mathbf y}}}
\def\bys{{\mathbf y}^*}

\def\dataset{\mathcal{D}}

\def\int{\mathrm{int}}

\newcommand{\valnet}[0]{v}
\newcommand{\valnets}[0]{v^*}

\newcommand{\byith}[1]{\by^{(#1)}}

\newcommand{\loss}[2]{\ell(#1, #2)}

\newcommand{\trans}[1]{{#1}^{\ensuremath{\mathsf{T}}}}

\def\eg{{\em e.g.,}}
\def\cf{{\em cf. }}
\def\ie{{\em i.e.,}}

\def\vs{{\em vs.}\xspace}

\newcommand{\tabref}[1]{Table~\ref{#1}}
\newcommand{\figref}[1]{Figure~\ref{#1}}
\newcommand{\secref}[1]{Section~\ref{#1}}

\newcommand{\WRP}{\par\qquad\qquad\enspace}

\icmltitlerunning{Deep Value Networks}

\begin{document} 

\twocolumn[
\icmltitle{Deep Value Networks Learn to\\
           Evaluate and Iteratively Refine Structured Outputs}



\icmlsetsymbol{intern}{*}

\begin{icmlauthorlist}
\icmlauthor{Michael Gygli}{eth,intern}
\icmlauthor{~~Mohammad Norouzi}{goo}
\icmlauthor{~~Anelia Angelova}{goo}
\end{icmlauthorlist}

\icmlaffiliation{eth}{ETH Z{\"u}rich \& gifs.com}
\icmlaffiliation{goo}{Google Brain, Mountain View, USA}

\icmlcorrespondingauthor{Michael Gygli}{gygli@vision.ee.ethz.ch}
\icmlcorrespondingauthor{Mohammad Norouzi}{mnorouzi@google.com}


\icmlkeywords{deep value network, structured prediction, semantic segmentation}

\vskip 0.3in
]



\printAffiliationsAndNotice{\doneasintern}  

\begin{abstract}
We approach structured output prediction by optimizing a {\em deep
  value network} (DVN) to precisely estimate the task loss on
different output configurations for a given input.  Once the model is
trained, we perform inference by gradient descent on the continuous
relaxations of the output variables to find outputs with promising
scores from the value network. When applied to image segmentation, the
value network takes an image and a segmentation mask as inputs and
predicts a scalar estimating the intersection over union between the
input and ground truth masks. For multi-label classification, the
DVN's objective is to correctly predict the F1 score for any potential
label configuration. The DVN framework achieves the state-of-the-art
results on multi-label prediction and image segmentation benchmarks.
\end{abstract}

\section{Introduction}

Structured output prediction is a fundamental problem in machine
learning that entails learning a mapping from input objects to complex
multivariate output structures. Because structured outputs live in a
high-dimensional combinatorial space, one needs to design factored
prediction models that are not only {\em expressive}, but also {\em
  computationally tractable} for both learning and inference. Due to
computational considerations, a large body of previous work
(\eg~\citet{crf, tsochantaridishja04}) has focused on relatively weak
graphical models with pairwise or small clique potentials. Such models
are not capable of learning complex correlations among the random
variables, making them not suitable for tasks requiring complicated
high level reasoning to resolve ambiguity.

\comment{
Recent work has applied expressive neural networks to structured
prediction to achieve impressive results on machine
translation~\cite{seq2seq,attention} and image and audio
synthesis~\cite{pixelcnn,wavenet,dahl2017pixel}. Such auto-regressive models
impose an order on the output variables and predict outputs one
variable at a time by formulating a locally normalized
probabilistic model. While training is often efficient, the key
limitation of such models is inference complexity, which grows
linearly in the number of output dimensions.
}

An expressive family of energy-based models studied by
\citet{lecun2006tutorial} and \citet{belanger2015structured} exploits
a neural network to score different joint configurations of inputs and
outputs. Once the network is trained, one simply resorts to
gradient-based inference as a mechanism to find low energy
outputs. Despite recent developments, optimizing parameters of deep
energy-based models remains challenging, limiting their
applicability. Moving beyond large margin training used by
previous work~\cite{belanger2015structured}, this paper presents a
simpler and more effective objective inspired by value based
reinforcement learning for training energy-based models.

Our key intuition is that learning to {\em critique} different output
configurations is easier than learning to directly come up with
optimal predictions. Accordingly, we build a deep value network (DVN)
that takes an input $\bx$ and a corresponding output structure $\by$,
both as inputs, and predicts a scalar score $v(\bx, \by)$ evaluating
the quality of the configuration $\by$ and its correspondence with the
input $\bx$. We exploit a loss function $\ell(\by, \bys)$ that
compares an output $\by$ against a ground truth label $\bys$ to teach
a DVN to evaluate different output configurations. The goal is to
distill the knowledge of the loss function into the weights of a value
network so that during inference, in the absence of the labeled output
$\bys$, one can still rely on the value judgments of the neural net to
compare outputs.

\begin{figure}[t]
\small
\begin{tabular}{@{}c@{\hspace*{.15cm}}c@{\hspace*{.0cm}}c@{\hspace*{.0cm}}c@{
\hspace*{.0cm}}@{\hspace*{.05cm}}c@{}} \\ &
  \multicolumn{3}{c@{\hspace*{.0cm}}@{\hspace*{.06cm}}}{Gradient based inference} & \\[.1cm]
  Input $\bx$ & Step 5 & Step 10 & Step 30 & GT label $\bys$\\
  \includegraphics[width=0.195\linewidth]{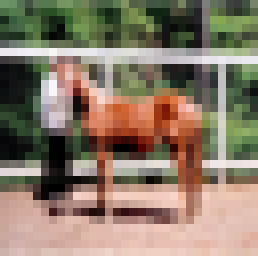} &
  \includegraphics[width=0.195\linewidth]{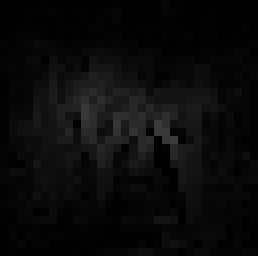} &
  \includegraphics[width=0.195\linewidth]{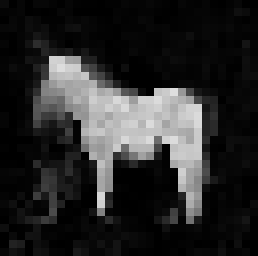} &
  \includegraphics[width=0.195\linewidth]{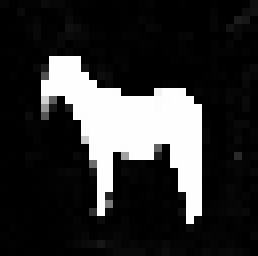} &
  \includegraphics[width=0.195\linewidth]{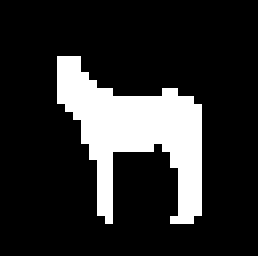} 
  \\[-.1cm]
  \includegraphics[width=0.195\linewidth]{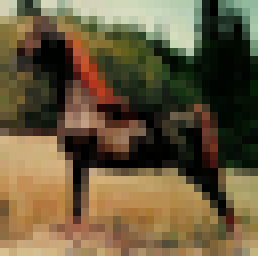} &
  \includegraphics[width=0.195\linewidth]{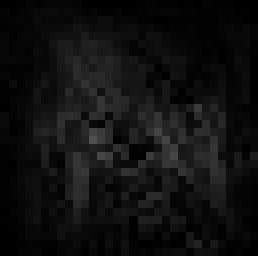} &
  \includegraphics[width=0.195\linewidth]{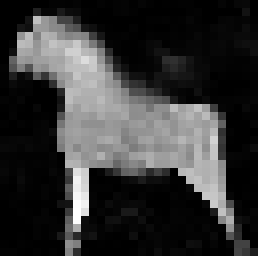} &
  \includegraphics[width=0.195\linewidth]{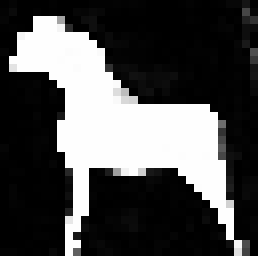} &
  \includegraphics[width=0.195\linewidth]{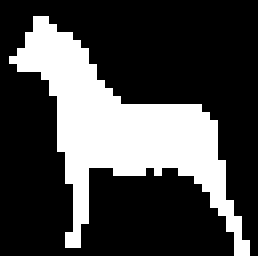} 
  \\[-.1cm]
  \includegraphics[width=0.195\linewidth]{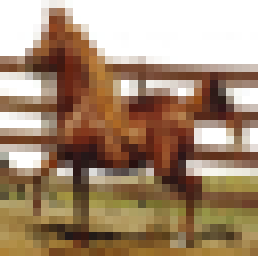} &
  \includegraphics[width=0.195\linewidth]{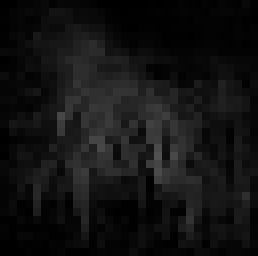} &
  \includegraphics[width=0.195\linewidth]{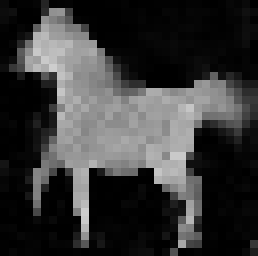} &
  \includegraphics[width=0.195\linewidth]{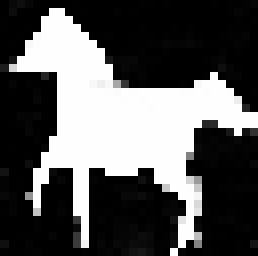} &
  \includegraphics[width=0.195\linewidth]{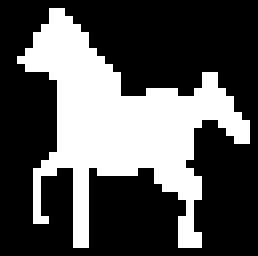} 
  \\[-.1cm]
  \includegraphics[width=0.195\linewidth]{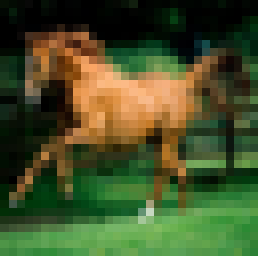} &
  \includegraphics[width=0.195\linewidth]{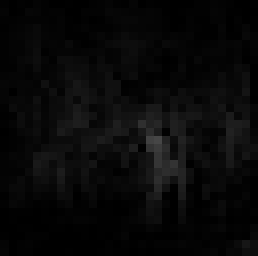} &
  \includegraphics[width=0.195\linewidth]{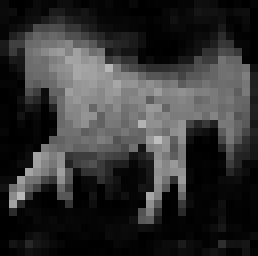} &
  \includegraphics[width=0.195\linewidth]{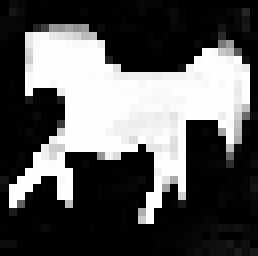} &
  \includegraphics[width=0.195\linewidth]{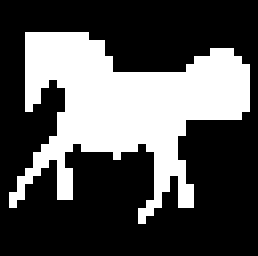} 
  \\[-.1cm]
\end{tabular}
\caption{Segmentation results of DVN on Weizmann horses test samples. Our
  gradient based inference method iteratively refines segmentation
  masks to maximize the predicted scores of a deep value network. Starting from a black mask at step $0$, the predictions
  converge within $30$ steps yielding the output segmentation.
    See \url{https://goo.gl/8OLufh} for more \& animated results.}
\label{fig:deap-dream-examples}
\end{figure}

To enable effective iterative refinement of structured outputs via
gradient ascent on the score of a DVN, similar to
\citet{belanger2015structured}, we relax the discrete output variables
to live in a continuous space. Moreover, we extend the domain of loss
functions so the loss applies to continuous variable outputs. For
example, for multi-label classification, instead of enforcing each
output dimension $y_i$ to be binary, we let $y_i \in [0, 1]$ and we
generalize the notion of $F_1$ score to apply to continuous
predictions. For image segmentation, we use a similar
generalization of intersection over union.
Then, we train a DVN on many output examples encouraging the network
to predict precise (negative) loss scores for almost any output
configuration. \figref{fig:deap-dream-examples} illustrates the
gradient based inference process on a DVN optimized for image
segmentation.

This paper presents a novel training objective for deep structured output
prediction, inspired by value-based reinforcement learning algorithms,
to precisely evaluate the quality of any input-output pair. We assess the
effectiveness of the proposed algorithm on multi-label classification
based on text data and on image segmentation.
We obtain state-of-the-art results in both cases, despite the differences of the domains and loss functions.
Even given a small number of input-output
pairs, we find that we are able to build powerful structure prediction
models. For example, on the Weizmann horses
dataset~\cite{borenstein2004learning}, without any form of
pre-training, we are able to optimize $2.5$ million network parameters
on only $200$ training images with multiple crops. Our deep value
network setup outperforms methods that are {\it pre-trained} on large
datasets such as ImageNet~\cite{deng2009imagenet} and methods that
operate on $4\times$ larger inputs. Our source code based on TensorFlow~\cite{tf} is available at
\url{https://github.com/gyglim/dvn}.

\section{Background}

Structured output prediction entails learning a mapping from input
objects $\bx \in \calX$ (\eg~$\calX \equiv \Real^M$) to multivariate
discrete outputs $\by \in \calY$ (\eg~$\calY \equiv \{0,
1\}^N$). Given a training dataset of input-output pairs, $\dataset
\equiv \{(\bx^{(i)}, \by^{*(i)})\}_{i=1}^N$, we aim to learn a mapping
$\byh(\bx): \calX \to \calY$ from inputs to ground truth
outputs. Because finding the exact ground truth output structures in a
high-dimensional space is often infeasible, one measures the quality
of a mapping via a loss function $\ell(\by, \by'): \calY \times \calY
\to \Real^+$ that evaluates the distance between different output
structures. Given such a loss function, the quality of a mapping is
measured by empirical loss over a validation dataset $\dataset'$,
\begin{equation}
\sum_{(\bx,\bys) \in \dataset'}
\loss{\byh(\bx)}{\bys}
\label{eq:empirical-loss}
\end{equation}
This loss can take an arbitrary form and is often non-differentiable.
For multi-label classification, a common loss is negative $F_1$ score
and for image segmentation, a typical loss is negative intersection
over union (IOU).

Some structured output prediction
methods~\cite{taskargk03,tsochantaridishja04} learn a mapping from
inputs to outputs via a score function $s(\bx, \by; \btheta)$, which
evaluates different input-output configurations  based
on a linear function of some joint input-output features $\psi(\bx,
\by)$,
\begin{equation}
  s(\bx, \by; \btheta) = \trans{\btheta} \psi(\bx, \by)\:.
\label{eq:linear-score}
\end{equation}
The goal of learning is to optimize a score function such that the
model's predictions denoted $\byh$,
\begin{equation}
\byh ~=~ \argmax{\by}{~s(\bx,\by; \btheta)} \:,
\label{eq:ssvm_argmax}
\end{equation} 
are closely aligned with ground-truth labels $\by^*$ as measured by
empirical loss in \eqref{eq:empirical-loss} on the training set.

Empirical loss is not amenable to numerical optimization because the
$\mathrm{argmax}$ in \eqref{eq:ssvm_argmax} is
discontinuous. Structural SVM
formulations~\cite{taskargk03,tsochantaridishja04} introduce a margin
violation (slack) variable for each training pair, and define a
continuous upper bound on the empirical loss.  The upper bound on the loss for
an example $(\bx, \bys)$ and the model's prediction $\widehat{\by}\, $
takes the form:%
\begin{subequations}
\begin{align}
\lefteqn{\hspace*{-.3cm}\ell(\widehat{\by}, \by^*)}
\nonumber \\ 
&\le ~ \max_{\by} \left[\, \ell({\by}, \by^*) \!+\!  s(\bx,\by; \btheta)
  \,\right] - s(\bx,\widehat{\by}; \btheta) \label{eq:loss-bound1} \\
&\le ~ \max_{\by}\left[\, \ell({\by}, \by^*) + s(\bx,\by; \btheta)
  \,\right] - s(\bx,{\by^*}; \btheta) ~.
\label{eq:loss-bound2}
\end{align}
\end{subequations}
Previous work~\cite{taskargk03,tsochantaridishja04}, defines a
surrogate objective on the empirical loss, by summing over the bound in
\eqref{eq:loss-bound2} for different training examples, plus a
regularizer. This surrogate objective is convex in $\btheta$, which
makes optimization convenient.

This paper is inspired by the structural SVM formulation above, but we
give up the convexity of the objective to obtain more expressive models
using a multi-layer neural networks. Specifically, we generalize the
formulation above in three ways: 1)~use a non-linear score function
denoted $v(\bx,\by; \btheta)$ that fuses $\psi(\cdot,\cdot)$ and
$\btheta$ together and jointly learns the features. 2)~use gradient descend in
$\by$ for iterative refinement of outputs to approximately find the
best $\byh(\bx)$. 3)~optimize the score function with a regression
objective so that the predicted scores closely approximate the
negative loss values,
\begin{equation}
  \forall \by \in \calY,~~~v(\bx,\by; \btheta) \approx -\ell(\by,
  \bys)~.
\label{eq:value-estimate}
\end{equation}
Our deep value network (DVN) is a non-linear function trying to
evaluate the value of any output configuration $\by \in \calY$
accurately. In the structural SVM's objective, the score surface can
vary as long as it does not violate margin constraints in
\eqref{eq:loss-bound2}. By contrast, we restrict the score surface
much more by penalizing it whenever it over- or underestimates the
loss values. This seems to be beneficial as a neural network
$v(\bx,\by; \btheta)$ has a lot of flexibility, and adding more
suitable constraints can help regularization.

We call our model a {\em deep value network (DVN)} to emphasize the
importance of the notion of {\em value} in shaping our ideas, but the
DVN architecture can be thought as an example of structured prediction
energy network (SPEN)~\cite{belanger2015structured} with similar
inference strategy. Belanger \& McCallum rely on the structural SVM
surrogate objective to train their SPENs, whereas inspired by
value based reinforcement learning, we learn an accurate estimate of
the values as in~\eqref{eq:value-estimate}. Empirically, we find that
the DVN outperforms large margin SPENs on multi-label classification
using a similar neural network architecture.

\section{Learning a Deep Value Network}
We propose a deep value network architecture, denoted $\valnet(\bx,
\by; \btheta)$, to evaluate a joint configuration of an input and a
corresponding output via a neural network. More specifically, the deep
value network takes as input both $\bx$ and $\by$ jointly, and after
several layers followed by non-linearities, predicts a scalar
$\valnet(\bx, \by; \btheta)$, which evaluates the quality of an output
$\by$ and its compatibility with $\bx$.  We assume that during
training, one has access to an oracle value function $\valnets(\by,
\bys) = -\ell(\by, \bys)$, which quantifies the quality of any $\by$. Such an oracle value function assigns optimal values to any
input-output pairs given ground truth labels $\bys$. During training,
the goal is to optimize the parameters of a value network, denoted
$\btheta$, to mimic the behavior of the oracle value function
$\valnets(\by, \bys)$ as much as possible.

Example oracle value functions for image segmentation and multi-label
classification include IOU and $F_1$ metrics, which are both
defined on $(\by, \bys) \in \{0, 1\}^M \times \{0, 1\}^M$,
\begin{equation}
  \valnets_{\text{IOU}}(\by, \bys) = \frac{\by \cap \bys}{\by \cup \bys}~,
  \label{eq:iou}
\end{equation}
\begin{equation}
  \valnets_{F_1}(\by, \bys) = \frac
  {2\,(\by \cap \bys)}
  {(\by \cap \bys) + (\by \cup \bys)}~.
  \label{eq:f1}
\end{equation}
Here $\by \cap \bys$ denotes the number of dimension $i$ where both
$\yi$ and $\yis$ are active and $\by \cup \bys$ denotes the number of
dimensions where at least one of $\yi$ and $\yis$ is active. Assuming
that one has learned a suitable value network that attains
$\valnet(\bx, \by; \btheta) \approx \valnets(\by, \bys)$ at every
input-output pairs, in order to infer a prediction for an input $\bx$,
which is valued highly by the value network, one needs to find $\byh =
\mathrm{argmax}_{\by}{~v(\bx,\by; \btheta)}$ as described below.

\subsection{Gradient based inference}
\label{sec:grad-inf}

Since $v(\bx,\by; \btheta)$ represents a complex non-linear function
of $(\bx, \by)$ induced by a neural network, finding $\byh$ is not
straightforward, and approximate inference algorithms based on
graph-cut~\cite{boykov2001fast} or loopy belief
propagation~\cite{murphy1999loopy} are not easily applicable. Instead,
we advocate using a simple gradient descent optimizer for inference.
To facilitate that, we relax the structured output variables to live
in a real-valued space. For example, instead of using $\by \in \{0,
1\}^M$, we use $\by \in [0, 1]^M$. The key to make this inference
algorithm work is that during training we make sure that our value
estimates are optimized along the inference trajectory.
Alternatively, one can make use of input convex neural
networks~\cite{amos2016input} to guarantee convergence to optimal $\widehat{\by}$.

Given a continuous variable $\by$, to find a local optimum of
$\valnet(\bx, \by; \btheta)$ {\em w.r.t.} $\by$, we start from an
initial prediction $\byith{0}$ (\ie~$\byith{0}=[0]^M$ in all of our
experiments), followed by gradient ascent for several steps,
\begin{equation}
  \byith{t + 1} = \calP_\calY \Big( \byith{t} + \eta \, \frac{\deriv}{\deriv \by}
  \, \valnet(\bx, \byith{t}; \btheta) \Big)~,
\label{eq:deepdream}
\end{equation}
where $\calP_\calY$ denotes an operator that projects the predicted
outputs back to the feasible set of solutions so that $\byith{t + 1}$
remains in $\calY$. In the simplest case, where $\calY = [0, 1]^M$,
the $\calP_\calY$ operator projects dimensions smaller than zero back
to zero, and dimensions larger than one to one. After the final
gradient step $T$, we simply round $\byith{T}$ to become discrete.
Empirically, we find that for a trained DVN, the generated
$\byith{T}$'s tend to become nearly binary themselves.

\subsection{Optimization}

To train a DVN using an oracle value function, first, one needs to
extend the domain of $v^*(\by, \bys)$ so it applies to continuous
output $\by$'s. For our IOU and $F_1$ scores, we simply extend the
notions of intersection and union by using element-wise min and max
operators,
\begin{eqnarray}
  \label{eq:real-cap} 
  \by \cap \bys &=&  \sum\nolimits_{i=1}^M \min \left(\yi, \yis \right)~,
  \\
  \by \cup \bys &=& \sum\nolimits_{i=1}^M \max \left(\yi, \yis \right)~.
  \label{eq:real-cup}
\end{eqnarray}
Substituting \eqref{eq:real-cap} and \eqref{eq:real-cup} into
\eqref{eq:iou} and \eqref{eq:f1} provides a generalization of IOU and
$F_1$ score to $[0, 1]^M \times [0, 1]^M$.

Our training objective aims at minimizing the discrepancy between
$v(\bx^{(i)}, \by^{(i)})$ and $v^{*(i)}$ on a training set of triplets
(input, output, value$^*$) denoted $\dataset \equiv \{(\bx^{(i)},
\by^{(i)}, v^{*(i)}\}_{i=1}^N$. Very much like Q-learning~\cite{watkins1992qlearning}, this
training set evolves over time, and one can make use of an experience
replay buffer.
In \secref{sec:triplet}, we discuss several strategies to
generate training tuples and in our experiments we evaluate such
strategies in terms of their empirical loss, once a gradient based
optimizer is used to find $\byh$.

Given a dataset of training tuples, one can use an appropriate loss to regress
$v(\bx, \by)$ to $v^*$ values. 
More specifically,
since both IOU and $F_1$ scores lie between $0$ and $1$, we used a
cross-entropy loss between oracle values \vs~our DVN values. As such,
our neural network $v(\bx, \by)$ has a sigmoid non-linearity at the
top to predict a number between $0$ and $1$, and the loss takes the
form,
\begin{equation}
\begin{aligned}
\mathcal{L}_{\mathrm{CE}}(\btheta) = \!\!\sum_{(\bx, \by, v^*)\in
  \dataset}\!\!\! &-v^* \log \valnet(\bx, \by; \btheta) \\
&- (1 - v^*) \log (1 - \valnet(\bx, \by; \btheta))
\end{aligned}
\label{eq:ce-loss}
\end{equation}
The exact form of the loss does not have a significant
impact on the performance and other loss functions can be used, e.g., $L_2$.  
A high level overview for training a DVN is shown in Algorithm~\ref{algo:training}.
For simplicity, we show the case when not using a queue and batch size $=1$.

\subsection{Generating training tuples}
\label{sec:triplet}
Each training tuple comprises an input, an output, and a corresponding
oracle value, \ie~$(\bx, \by, v^{*})$. The way training tuples are
generated significantly impacts the performance of our structured
prediction algorithm. In particular, it is important that the tuples
are chosen such that they provide a good coverage of the space of
possible outputs and result in a large learning signal. There exist
several ways to generate training tuples including:
\vspace{-.4cm}
\begin{itemize}
  \setlength{\parskip}{0pt}
  \setlength{\itemsep}{0pt plus 1pt}
\item running gradient based {\em inference} during training.
\item generating {\em adversarial tuples} that have a large
  discrepancy between $\valnet(\bx, \by; \btheta)$ and $v^*(\by,
  \bys)$.
\item {\em random samples} from $\calY$, maybe biased towards $\bys$.
\end{itemize}%
\vspace{-.4cm}
We elaborate on these methods below, and present a comparison of their
performance in~\secref{sec:ablation}. Our ablation
experiments suggest that combining examples from gradient based inference
with adversarial tuples works best.

\noindent
\textbf{Ground truth.}
In this setup we simply add the ground truth outputs $\bys$ into training
with a $v^* = 1$ to provide some positive examples.

\noindent
\textbf{Inference.} In this scenario, we generate samples by running a
gradient based inference algorithm (\secref{sec:grad-inf}) along our
training. This procedure is useful because it helps learning a good
value estimate on the output hypotheses that are generated along the
inference trajectory at test time. To speed up training, we run a
parallel inference job using slightly older neural network weights and
accumulate the inferred examples in a queue.

\noindent
\textbf{Random samples.}  In this approach, we sample a solution $\by$
proportional to its exponentiated oracle value, \ie~$\by$ is sampled
with probability $p(\by) \propto \exp\{\valnets(\by, \bys)/\tau\}$,
where $\tau > 0$ controls the concentration of samples in the
vicinity of the ground truth. At $\tau = 0$ we recover the ground
truth samples above. We follow~\cite{norouzi2016reward} and sample
from the exponentiated value distribution using stratified sampling,
where we group $\by$'s according to their values. This approach
provides a good coverage of the space of possible solutions.

\noindent
\textbf{Adversarial tuples.}  We maximize the cross-entropy loss used
to train the value network~\eqref{eq:ce-loss} to generate adversarial
tuples again using a gradient based optimizer
(\eg~see~\cite{goodfellow2014explaining,szegedy2013intriguing}. Such
adversarial tuples are the outputs $\by$ for which the network over-
or underestimates the oracle values the most. This strategy finds some
difficult tuples that provide a useful learning signal, while ensuring
that the value network has a minimum level of accuracy across all
outputs $\by$.

\begin{algorithm}[t]
\small
\begin{algorithmic}[1]
\Function{TrainEpoch}{training buffer $\dataset$, initial weights $\btheta$, learning rate $\lambda$} 
  \While{not converged}
  \State $(\bx,\bys) \thicksim \dataset$  \Comment{Get a training example}
 \State $\by \gets$ \Call{GenerateOuput}{$\bx,\btheta$} \Comment{\cf Sec.~\ref{sec:triplet}} 
 \State $v^* \gets \valnets(\by, \bys) $ \Comment{Get oracle value for $\by$} 
 \State \(\triangleright\) Compute loss based on estimation error \cf \eqref{eq:ce-loss}
 \State $\mathcal{L} \gets -v^* \log \valnet(\bx, \by; \btheta)$ \WRP $- (1 - v^*) \log (1 - \valnet(\bx, \by; \btheta))$
 \State $\btheta \gets \btheta - \lambda \frac{\deriv}{\deriv \btheta} \mathcal{L}$   \Comment{Update DVN weights}
 \EndWhile
\EndFunction
\end{algorithmic}
 \caption{Deep Value Network training}
 \label{algo:training}
\end{algorithm}

\section{Related work}

There has been a surge of recent interest in using neural networks for
structured prediction~\cite{zheng2015crfnet,
  chen2015learning,song2016training}.  The Structured Prediction
Energy Network (SPEN) of~\cite{belanger2015structured} inspired in
part by~\cite{lecun2006tutorial} is identical to the DVN
architecture. Importantly, the motivation and the learning objective
for SPENs and DVNs are distinct -- SPENs rely on a max-margin
surrogate objective whereas we directly regress the energy of an
input-output pair to its corresponding loss. Unlike SPENs that only
consider multi-label classification problems, this has allowed us to train a deep
convolutional network to successfully address complex image
segmentation problems.
Concurrent to our work, \cite{belanger2017end} explored another way of improving the training of SPENs, by directly back-propagating the error through the gradient-based inference process. This requires expensive gradient computation via unrolling of the computation graph for the number of inference gradient steps. By contrast, our training algorithm is much more efficient only requiring back-propagation through the value network once.

Recent work has applied expressive neural networks to structured
prediction to achieve impressive results on machine
translation~\cite{seq2seq,attention} and image and audio
synthesis~\cite{pixelcnn,wavenet,dahl2017pixel}. Such autoregressive
models impose an order on the output variables and predict outputs one
variable at a time by formulating a locally normalized probabilistic
model. While training is often efficient, the key limitation of such
models is inference complexity, which grows linearly in the number of
output dimensions; this is not acceptable for high-dimensional output
structures. By contrast, inference under our method is efficient as
all of the output dimensions are updated in parallel.

Our approach is inspired in part by the success of previous work on
value-based reinforcement learning (RL) such as
Q-learning~\cite{qlearning,watkins1992qlearning}
(see~\cite{sutton98reinforcement} for an overview). The main idea is
to learn an estimate of the future reward under the optimal behavior
policy at any point in time. Recent RL algorithms use a neural network
function approximator as the model to estimate the action
values~\cite{dqn}. We adopt similar ideas for structured output
prediction, where we use the task loss as the optimal value estimate.
Unlike RL, we use a gradient based inference algorithm to find
optimal solutions at test time.

Gradient based inference, sometimes called {\em deep dreaming} has led
to impressive artwork and has been influential in designing
DVN~\cite{gatys2015neural,mordvintsev2015inceptionism,
  nguyen2016synthesizing,dumoulin2016learned}. Deep dreaming and style
transfer methods iteratively refine the input to a neural net to
optimize a prespecified objective. Such methods often use a {\em
  pre-trained} network to define a notion of a perceptual
loss~\cite{johnson2016perceptual}. By contrast, we train a task
specific value network to learn the characteristics of a task specific
loss function and we {\em learn} the network's weights from scratch.

Image segmentation~\cite{arbelaez2012semantic,
  carreira2012semantic,girshick2014rich,hariharan2015hypercolumns}, is
a key problem in computer vision and a canonical example of structured
prediction.  Many segmentation approaches based on Convolutional
Neural Networks (CNN) have been
proposed~\cite{girshick2014rich,chen2014semantic,eigen2015predicting,long2015fully,
  ronneberger2015unet,noh2015learning}. Most use a deep neural network
to make a per-pixel prediction, thereby modeling pairs of pixels as
being conditionally independent given the input.

To diminish the conditional independence problem, recent techniques
propose to model dependencies among output labels to refine an initial
CNN-based coarse segmentation.  Different ways to incorporate pairwise
dependencies within a segmentation mask to obtain more expressive
models are proposed in
\cite{chen2014semantic,chen2016deeplab,ladicky2013inference,zheng2015crfnet}.
Such methods perform joint inference of the segmentation mask
dimensions via graph-cut~\cite{li2015object}, message
passing~\cite{koltun2011efficient} or loopy belief
propagation~\cite{murphy1999loopy}, to name a few variants. Some
methods incorporate higher order potentials in
CRFs~\cite{kohli2009robust} or model global shape priors with
Restricted Boltzmann
Machines~\cite{li2013exploring,kae2013augmenting,yang2014max,eslami2014shape}.
Other methods learn to iteratively refine an initial prediction by
CNNs, which may just be a coarse segmentation
mask~\cite{safar2015learning,pinheiro2016learning,li2016iterative}.

By contrast, this paper presents a new framework for training a score
function by having a gradient based inference algorithm in mind during
training. Our deep value network applies to generic structured
prediction tasks, as opposed to some of the methods above, which
exploit complex combinatorial structures and special constraints such
as sub-modularity to design inference algorithms. Rather, we use
expressive energy models and the simplest conceivable inference
algorithm of all -- gradient descent.

\section{Experimental evaluation}
\label{sec:exp}

We evaluate the proposed Deep Value Networks on $3$ tasks: multi-label
classification, binary image segmentation, and a $3$-class face
segmentation task.  \secref{sec:ablation} investigates the sampling
mechanisms for DVN training, and \secref{sec:vis} visualizes the
learned models.

\subsection{Multi-label classification}
\label{sec:tag}
We start by evaluating the method on the task of predicting 
tags from text inputs. We use standard benchmarks in multi-label
classification, namely Bibtex and Bookmarks, introduced
in~\cite{katakis2008multilabel}. In this task, multiple labels are
possible per example, and the correct number is not known. Given the
structure in the label space, methods modeling label correlations
often outperform models with independent label predictions. We compare
DVN to standard baselines including per-label logistic
regression from~\cite{linmulti14}, and a two-layer neural network with
cross entropy loss~\cite{belanger2015structured}, as well as
SPENs~\cite{belanger2015structured} and PRLR~\cite{linmulti14}, which is the state-of-the-art on these datasets. To allow direct comparison with
SPENs, we adopt the same architecture in this paper. Such an
architecture combines local predictions that are non-linear in $\bx$,
but linear in $\by$, with a so-called global network, which scores
label configuration with a non-linear function of $\by$ independent of
$\bx$ (see~\citet{belanger2015structured}, Eqs. (3) - (5)). Both local
prediction and global networks have one or two hidden layers with Softplus
non-linerarities.  We follow the same experimental protocol and report
$F_1$ scores on the same test split as~\cite{belanger2015structured}.

\begin{table}[t]
\centering
\footnotesize
{\def\arraystretch{1.17}
\begin{tabular}{@{}|@{\hspace{.05cm}}l@{\hspace{.05cm}}|@{\hspace{.05cm}}c@{\hspace{.05cm}}|@{\hspace{.05cm}}c@{\hspace{.05cm}}|@{}}
\hline
Method   & Bibtex &Bookmarks \\
\hline
Logistic regression~\cite{linmulti14}                  & 37.2 & 30.7 \\
NN baseline~\cite{belanger2015structured}  & 38.9 & 33.8 \\
SPEN~\cite{belanger2015structured}                     & 42.2 & 34.4 \\
PRLR~\cite{linmulti14} & 44.2 & 34.9   \\
DVN (Ours) & 44.7 & 37.1 \\
\hline

\end{tabular}
\caption{Tag prediction from text data. $F_1$ performance of Deep Value
  Networks compared to the state-of-the-art on multi-label
  classification. All prior results are taken
  from~\cite{linmulti14, belanger2015structured}}
\label{mlc_results}
}
\end{table}

The results are summarized in~\tabref{mlc_results}.  As can be seen
from the table, our method outperforms the logistic regression
baselines by a large margin.  It also significantly improves over
SPEN, despite not using any pre-training. SPEN, on the other hand,
relies on pre-training of the feature network with a logistic loss to
obtain good results. Our results even outperform~\cite{linmulti14}. This is encouraging, as their method is specific to classification and encourages sparse and low-rank predictions, whereas our technique does not have such dataset specific regularizers.

\subsection{Weizmann horses}
\label{sec:weiz}
The Weizmann horses dataset~\cite{borenstein2004learning} is a dataset
commonly used for evaluating image segmentation
algorithms~\cite{li2013exploring,yang2014max,safar2015learning}.  The
dataset consists of $328$ images of left oriented horses and their
binary segmentation masks.  We
follow~\cite{li2013exploring,yang2014max,safar2015learning} and
evaluate the segmentation results at $32\!\times\!32$ dimensions.
Satisfactory segmentation of horses requires learning strong shape
priors and complex high level reasoning, especially at a low
resolution of $32\!\times\!32$ pixels, because small parts such as the
legs are often barely visible in the RGB image. We follow the
experimentation protocol of~\cite{li2013exploring} and report results
on the same test split.

\label{details}
\begin{figure}[t]
\includegraphics[width=\linewidth]{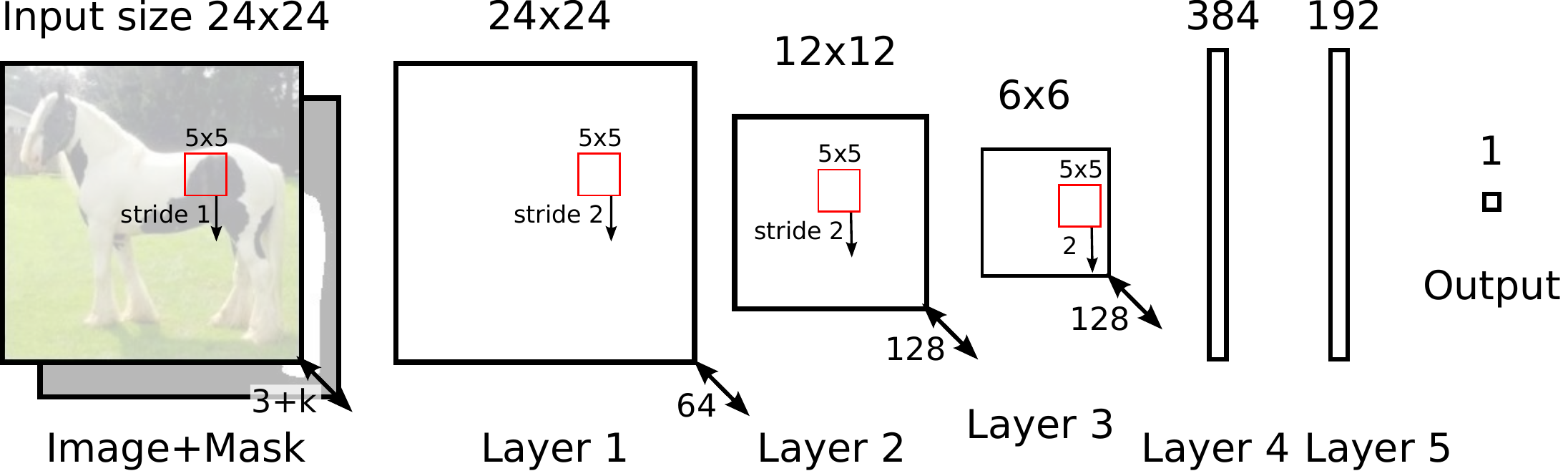}
\caption{A deep value network with a feed-forward convolutional architecture, used for segmentation. The network takes 
an image and a segmentation mask as input and predicts a scalar
evaluating the compatibility between the input pairs.}
\label{fig:arch}
\end{figure}

For the DVN we use a simple CNN architecture consisting
of 3 convolutional and 2 fully connected layers
(\figref{fig:arch}).  We use a learning rate of $0.01$ and apply
dropout on the first fully connected layer with the keeping
probability $0.75$ as determined on the validation set. We empirically
found $\tau=0.05$ to work best for stratified sampling.  For training
data augmentation purposes we randomly crop the image, similar
to~\cite{krizhevsky2012imagenet}.  At test time, various strategies
are possible to obtain a full resolution segmentation, which we
investigate in Section~\ref{sec:ablation}.  For comparison we also
implemented a Fully Convolutional Network (FCN)
baseline~\cite{long2015fully}, by using the same convolutional layers
as for the value network (\cf \figref{fig:arch}).  If not explicitly
stated, masks are averaged over over 36 crops for our model
and~\cite{long2015fully} (see below).

\begin{table}[t!]
\centering
\footnotesize
{\def\arraystretch{1.3}
\begin{tabular}{c@{\hspace*{.2cm}}c@{\hspace*{.2cm}}|l@{\hspace*{.1cm}}|@{\hspace*{.1cm}}c@{\hspace*{.1cm}}|@{\hspace*{.1cm}}c@{\hspace*{.1cm}}|}
\cline{3-5}
& & \textbf{Method}   & \textbf{Mean}   & \textbf{Global} \\
& &                   & \textbf{IOU \%} & \textbf{IOU \%} \\
\cline{2-5}
\parbox[t]{2mm}{\multirow{6}{*}{\rotatebox[origin=c]{90}{Input size}}} &
\parbox[t]{2mm}{\multirow{3}{*}{\rotatebox[origin=c]{90}{$32\!\times\!32$}}} &
CHOPPS~\cite{li2013exploring} & 69.9 &  - \\
& & Fully conv (FCN) baseline & 78.56 & 78.7 \\
& & DVN (Ours)    &  84.1 & 84.0 \\
\cline{2-5}
& \parbox[t]{2mm}{\multirow{3}{*}{\rotatebox[origin=c]{90}{$128\!\times\!128$}}} &
 MMBM2~\cite{yang2014max} &  - & 72.1 \\
& & MMBM2 + GC~\cite{yang2014max} & - & 75.8 \\
& & Shape NN~\cite{safar2015learning} & - & 83.5 \\
\cline{2-5}
\end{tabular}
}
\caption{Test IOU on Weizmann-$32\!\times\!32$ dataset. DVN outperforms all previous methods, despite using a much lower input resolution than~\cite{yang2014max} and \cite{safar2015learning}. }
\label{weizmann_results}
\end{table}

We test and compare our model on the Weizmann horses segmentation task
in \tabref{weizmann_results}.  We tune the hyper-parameters of the
model on a validation set and, once best hyper-parameters are found,
 fine-tune on the combination of training and validation sets.  We
report the mean image IOU, as well as the IOU over the whole test set, as commonly done in the literature.  It is clear that our
approach outperforms previous methods by a significant margin on both
metrics. Our model shows strong segmentation results, without relying
on externally trained CNN features
as~(\eg~\citet{safar2015learning}). The weights of our value network
are learned from scratch on crops of just $200$ training images.  Even
though the number of examples is very small for this dataset, we did
not observe overfitting during training, which we attribute to being
able to generate a large set of segmentation masks for training.

\begin{figure}[t]
  \begin{center}
    \begin{tabular}{@{}c@{\hspace*{.04cm}}c@{\hspace*{.04cm}}c@{\hspace*{.04cm}}c@{\hspace*{.04cm}}c@{\hspace*{.04cm}}c@{}}
     \small Input & \small CHOPPS [1] & \small FCN [2] & \small DVN & \small GT label
     \\
     \includegraphics[width=.19\linewidth]{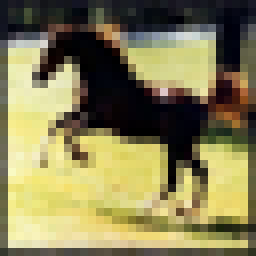} &
     \includegraphics[width=.19\linewidth]{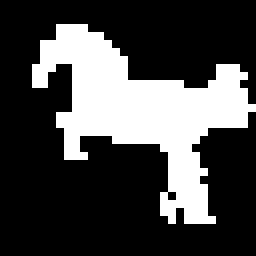} &
     \includegraphics[width=.19\linewidth]{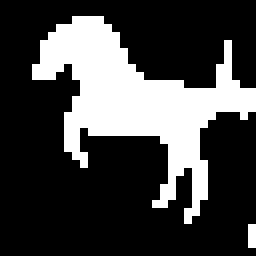} &
     \includegraphics[width=.19\linewidth]{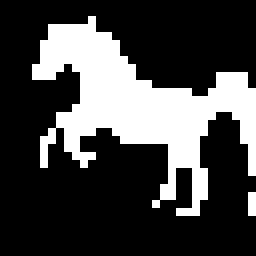} &
     \includegraphics[width=.19\linewidth]{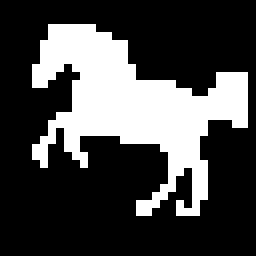}
     \\[-.1cm]
     \includegraphics[width=.19\linewidth]{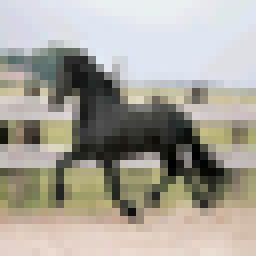} &
     \includegraphics[width=.19\linewidth]{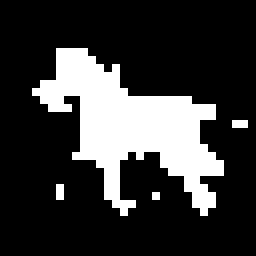} &
     \includegraphics[width=.19\linewidth]{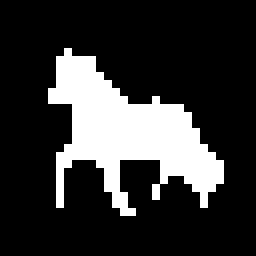} &
     \includegraphics[width=.19\linewidth]{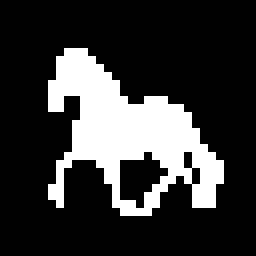} &
     \includegraphics[width=.19\linewidth]{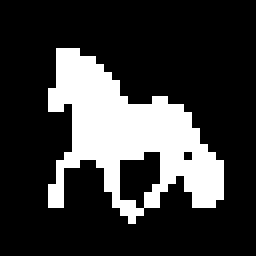}
     \\[-.1cm]
     \includegraphics[width=.19\linewidth]{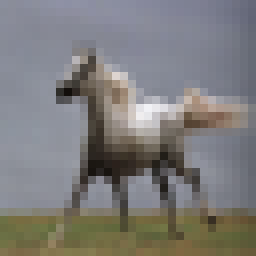} &
     \includegraphics[width=.19\linewidth]{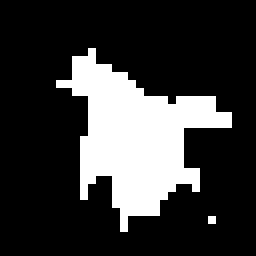} &
     \includegraphics[width=.19\linewidth]{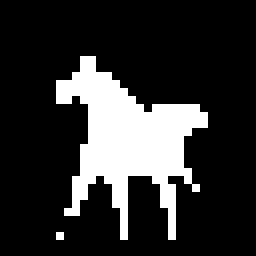} &
     \includegraphics[width=.19\linewidth]{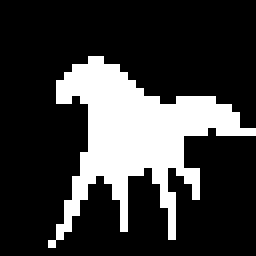} &
     \includegraphics[width=.19\linewidth]{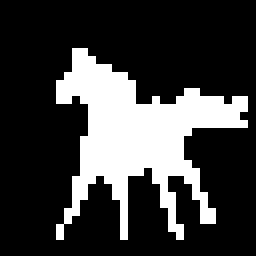}
     \\[-.1cm]
     \includegraphics[width=.19\linewidth]{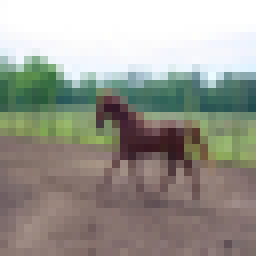} &
     \includegraphics[width=.19\linewidth]{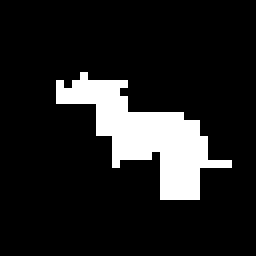} &
     \includegraphics[width=.19\linewidth]{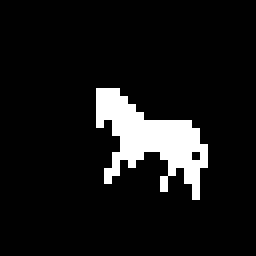} &
     \includegraphics[width=.19\linewidth]{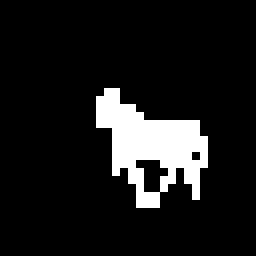} &
     \includegraphics[width=.19\linewidth]{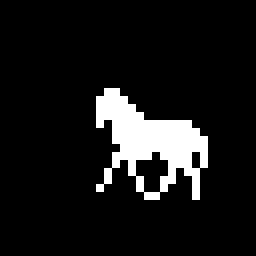}
     \\[-.1cm]
     \includegraphics[width=.19\linewidth]{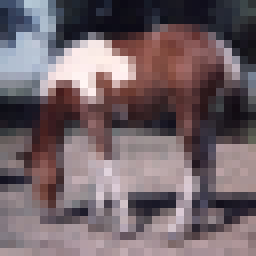} &
     \includegraphics[width=.19\linewidth]{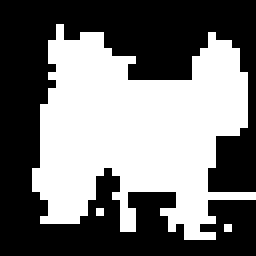} &
     \includegraphics[width=.19\linewidth]{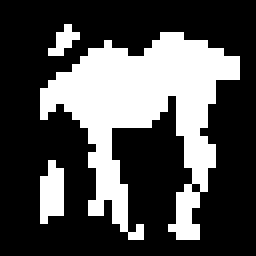} &
     \includegraphics[width=.19\linewidth]{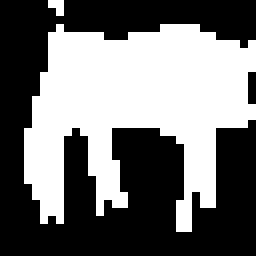} &
     \includegraphics[width=.19\linewidth]{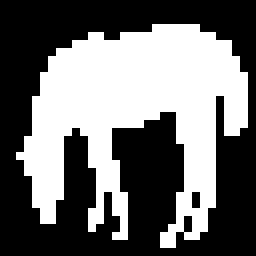}
     \\[-.1cm]
     \includegraphics[width=.19\linewidth]{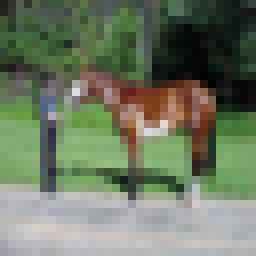} &
     \includegraphics[width=.19\linewidth]{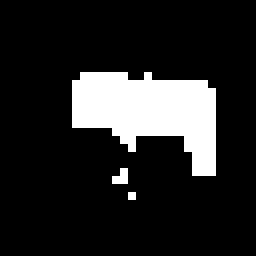} &
     \includegraphics[width=.19\linewidth]{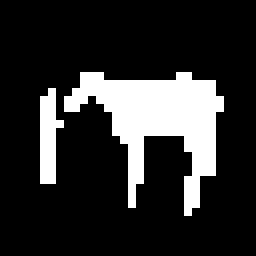} &
     \includegraphics[width=.19\linewidth]{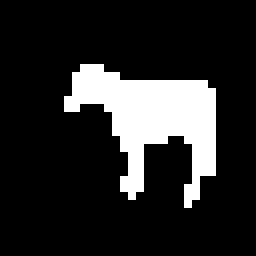} &
     \includegraphics[width=.19\linewidth]{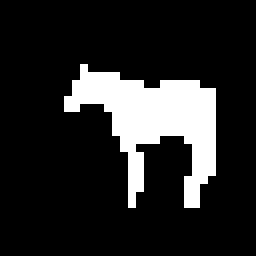}
     \\[-.1cm]
  \end{tabular}
  \end{center}
\caption{Qualitative results on the Weizmann $32\times32$ dataset. In
  comparison to previous works, DVN is able to learn a strong shape
  prior and thus correctly detect the horse shapes including
  legs. Previous methods are often misled by other objects or low
  contrast, thus generating inferior masks. References:
  [1]~\citet{li2013exploring}~~[2]~Our
  implementation of FCN~\cite{long2015fully}
}
\label{weizmann_examples}
\end{figure}

In \figref{weizmann_examples} we show qualitative results for CHOPPS~\cite{li2013exploring}, our
implementation of fully convolutional networks
(FCN)~\cite{long2015fully}, and our DVN model.  When comparing our model to FCN, trained
on the same data and resolution, we find that the FCN has challenges
correctly segmenting legs and ensuring that the segmentation masks
have a single connected component (\eg~\figref{weizmann_examples},
last two rows).  Indeed, the masks produced by the DVN correspond to
much more reasonable horse shapes as opposed to those of other methods
-- the DVN seem capable of learning complex shape models and
effectively grounding them to visual evidence. We also note that in
our comparison in \tabref{weizmann_results}, prior methods using
larger inputs (\eg~$128\!\times\!128$) are also outperformed by DVNs.

\subsection{Labeled Faces in the Wild}
\label{sec:faces}
The Labeled Faces in the Wild (LFW) dataset~\cite{huang2007labeled} was
proposed for face recognition and contains more than $13000$ images.
A subset of $2927$ faces was later annotated for segmentation
by~\citet{kae2013augmenting}. The labels are provided on a superpixel
basis and consist of $3$ classes: face, hair and background. We use
this dataset to test the application of our approach to multiclass
segmentation.  We use the same train, validation, and test splits
as~\cite{kae2013augmenting,tsogkas2015deep}.  As our method predicts
labels for pixels, we follow~\cite{tsogkas2015deep} and map pixel
labels to superpixels by using the most frequent label in a superpixel
as the class. To train the DVN, we use mean pixel accuracy as our
oracle value function, instead of superpixel accuracy.

\begin{figure}[t!]
  \begin{center}
    \begin{tabular}{@{}c@{\hspace*{.04cm}}c@{\hspace*{.04cm}}c@{\hspace*{.04cm}}}
      Input & DVN & GT label \\
         \includegraphics[width=17mm]{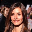} &
         \includegraphics[width=17mm]{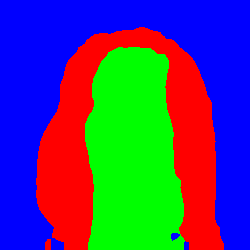}   &
         \includegraphics[width=17mm]{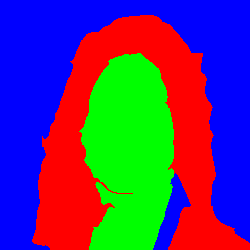}     \\[-.1cm]
         \includegraphics[width=17mm]{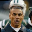} &
         \includegraphics[width=17mm]{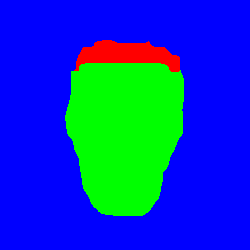}   &
         \includegraphics[width=17mm]{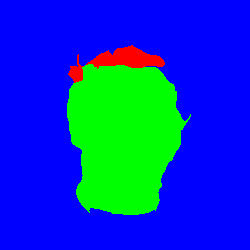}     \\[-.1cm]
         \includegraphics[width=17mm]{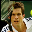} &
         \includegraphics[width=17mm]{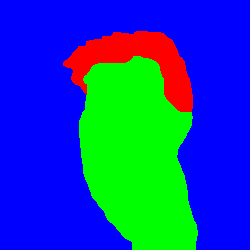}   &
         \includegraphics[width=17mm]{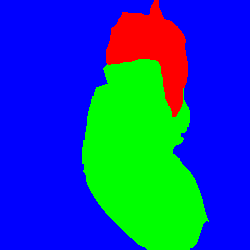}     \\[-.1cm]
         \includegraphics[width=17mm]{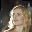} &
         \includegraphics[width=17mm]{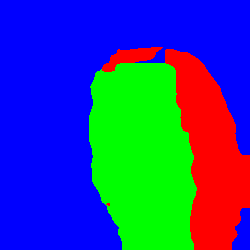}   &
         \includegraphics[width=17mm]{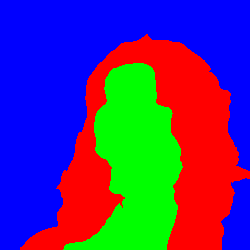}     \\[-.1cm]
         \includegraphics[width=17mm]{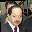} &
         \includegraphics[width=17mm]{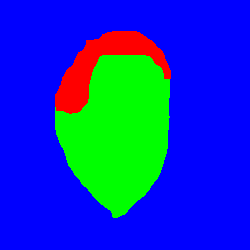}   &
         \includegraphics[width=17mm]{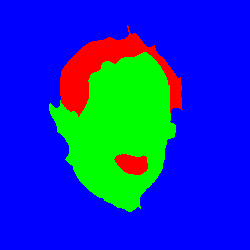}   \\         
  \end{tabular}
  \end{center}
    \caption{Qualitative results on $3$-class segmentation on the LFW
      dataset. The last two rows show failure cases, where our model
      does not detect some of hair and moustache correctly. }
\label{lfw_examples}
\end{figure}

Table~\ref{lfw_results} shows quantitative results. DVN
performs reasonably well, but is outperformed by state of the art
methods on this dataset.  We attribute this to three reasons. (i) the
pre-training and more direct optimization of the per-pixel prediction
methods of~\cite{tsogkas2015deep,long2015fully}, (ii) the input
resolution and (iii) the properties of the dataset.  In contrast to
 horses, faces do not have thin parts and exhibit limited
deformations. Thus, a feed forward method as used
in~\cite{long2015fully}, which produces coarser and smooth predictions
is sufficient to obtain good results. Indeed, this has also been
observed in the negligible improvement of refining CNN predictions
with Conditional Random Fields and Restricted Boltzmann machines (\cf
Table~\ref{lfw_results} last three rows).  Despite this, our model is
able to learn a prior on the shape and align it with the image
evidence in most cases.  Some failure cases include failing to
recognize subtle and more rare parts such as mustaches, given their
small size, and difficulties in correctly labeling blond hair.
Figure~\ref{lfw_examples} shows qualitative results of our
segmentation method on this dataset.

\begin{table}[t]
\centering
\footnotesize
{\def\arraystretch{1.17}
\begin{tabular}{c@{\hspace{.2cm}}c@{\hspace*{.3cm}}|l@{\hspace{.1cm}}|@{\hspace{.1cm}}c@{\hspace{.1cm}}|}
\cline{3-4}
& & \textbf{Method}   & \textbf{SP Acc. \%}  \\
\cline{2-4}
\parbox[t]{2mm}{\multirow{5}{*}{\rotatebox[origin=c]{90}{Input size}}} &
\parbox[t]{2mm}{\multirow{2}{*}{\rotatebox[origin=c]{90}{$32^2$}}} &
Fully conv (FCN) baseline & 95.36  \\
& & DVN (Ours)      & 92.44  \\
\cline{2-4}
& 
\parbox[t]{2mm}{\multirow{5}{*}{\rotatebox[origin=c]{90}{$250^2$}}} &
CRF (as in~\citet{kae2013augmenting}) & 93.23  \\
& & GLOC~\cite{kae2013augmenting} & 94.95   \\
& & DNN~\cite{tsogkas2015deep} & 96.54  \\
& & DNN+CRF+SBM~\cite{tsogkas2015deep} & 96.97  \\
\cline{2-4}
\end{tabular}
\caption{Superpixel accuracy (SP Acc.) on Labeled Faces in the Wild test set.}
\label{lfw_results}
}
\end{table}

\begin{table}[t]
\centering
\footnotesize
\begin{tabular}{|@{\hspace{.1cm}}l@{\hspace{.1cm}}|@{\hspace{.1cm}}c@{\hspace{.1cm}}|}
\hline
\textbf{Configuration}         & \textbf{Mean IOU \%}   \\
\hline
Inference + Ground Truth & 76.7\\
Inference + Stratified Sampling & 80.8 \\
Inference + Adversarial (DVN) & 81.6\\

\hline\hline
DVN + Mask averaging (9 crops) & 81.3 \\
DVN + Joint inference (9 crops) & 81.6 \\
\hline
DVN + Mask avg. non-binary (25 crops) & 69.6 \\
DVN + Joint inf. non-binary (25 crops) & 80.3 \\
\hline
DVN + Mask averaging (25 crops) & 83.1 \\
DVN + Joint inference (25 crops) & 83.1\\
\hline
\end{tabular}
\caption{Test performance of different configurations on the Weizmann 32x32 dataset.}
\label{weizmann_detailed_results}
\end{table}

\subsection{Ablation experiments}
\label{sec:ablation}
In this section we analyze different configurations of our method.  As
already mentioned, generating appropriate training data for our method
is key to learning good value networks. We compare $3$ main approaches: 1) inference +
ground truth, 2) inference + stratified sampling, and 3) inference +
adversarial training. These experiments are conducted on the Weizmann
dataset, described above. \tabref{weizmann_detailed_results}, top
portion, reports IOU results for different approaches for training the
dataset. As can be seen, including adversarial training works best,
followed by stratified sampling. Both of these methods help explore
the space of segmentation masks in the vicinity of ground truth masks
better, as opposed to just including the ground truth masks.  Adding
adversarial examples works better than stratified sampling, as the
adversarial examples are the masks on which the model is least
accurate.  Thus, these masks provide useful gradient information as to
help improve the model.

We also investigate ways to do model averaging
(Table~\ref{weizmann_detailed_results}, bottom portion).  Averaging
the segmentation masks of multiple crops leads to improved
performance.  When the masks are averaged na\"{i}vely, the result
becomes blurry, making it difficult to obtain a final segmentation.
Instead, joint inference updates the complete segmentation mask in
each step, using the gradients of the individual crops. This procedure
leads to clean, near-binary segmentation masks. This is manifested in
the performance when using the raw foreground confidence
(Table~\ref{weizmann_detailed_results}, Mask averaging non-binary \vs
Joint inference non-binary).  Joint inference leads to somewhat
improved segmentation results, even after binarization, in particular
when using fewer crops.

\begin{figure}[t]
\begin{subfigure}{1\linewidth}
    \centering
    \begin{subfigure}[t]{0.24\textwidth}    
  \includegraphics[width=\linewidth]{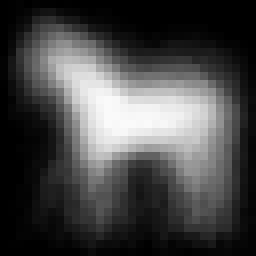}
  \caption{} 
    \end{subfigure}%
    ~
    \begin{subfigure}[t]{0.24\textwidth}    
  \includegraphics[width=\linewidth]{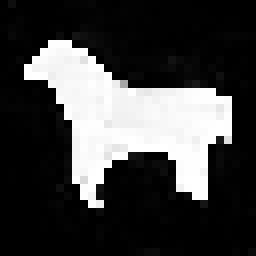}
  \caption{} 
    \end{subfigure}%
    ~
    \begin{subfigure}[t]{0.24\textwidth}    
  \includegraphics[width=\linewidth]{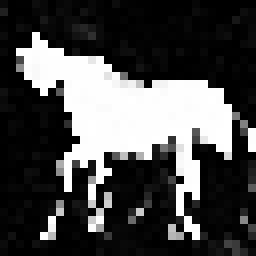} 
  \caption{} 
    \end{subfigure}%
    ~
    \begin{subfigure}[t]{0.24\textwidth}    
  \includegraphics[width=\linewidth]{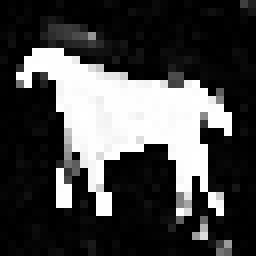}
  \caption{} 
    \end{subfigure}%
\end{subfigure}
\caption{Visualization of the learned horse shapes on the Weizmann
  dataset.  From left to right (a) The mean mask of the training set
  (b) mask generated when providing the mean horse image from the
  training set (c, d) Outputs generated by our model given mean horse
  image plus Gaussian noise ($\sigma=10$) as the input.}
\label{fig:prior}
\vspace*{-.2cm}
\end{figure}

\subsection{Visualizing the learned correlations}
\label{sec:vis}
To visualize what the model has learned, we run our inference
algorithm on the mean image of the Weizmann dataset (training
split). Optionally, we perturb the mean image by adding some Gaussian
noise. The masks obtained through this procedure are shown in
Figure~\ref{fig:prior}.  As one can see, the segmentation masks found
by the value network on (noisy) mean images resemble a side-view of a
horse with some uncertainty on the leg and head positions. These parts
have the most amount of variation in the dataset. Even though noisy
images do not contain horses, the value network hallucinates proper
horse silhouettes, which is what our model is trained on.

\section{Conclusion}

This paper presents a framework for structured output prediction by
learning a deep value network that predicts the
quality of different output hypotheses for a given input. As the DVN
learns to predict a value based on both, input and output, it
implicitly learns a prior over output variables and takes advantage of
the joint modelling of the inputs and outputs. By visualizing the
prior for image segmentation, we indeed find that our model learns
realistic shape priors.  Furthermore, rather than learning a model by
optimizing a surrogate loss, using DVNs allows to directly train a
network to accurately predict the desired performance metric (\eg~
IOU), even if it is non-differentiable. We apply our method to several
standard datasets in multi-label classification and image
segmentation. Our experiments show that DVNs apply to different
structured prediction problems, achieving state-of-the-art results with
no pre-training.

As future work, we plan to improve the scalability and computational
efficiency of our algorithm by inducing input features computed solely
on $\bx$, which is going to be computed only once. The gradient based
inference can improve by injecting noise to the gradient estimate,
similar to Hamiltonian Monte Carlo sampling. Finally, one can explore
better ways to initialize the inference process.

\section{Acknowledgment}

We thank Kevin Murphy, Ryan \& George Dahl, Vincent Vanhoucke,
Zhifeng Chen, and the Google Brain team for insightful comments and
discussions.

\bibliography{references}
\bibliographystyle{icml2017}

\end{document}


\title{Learning to Deep Dream Semantic Segmentations - Suplementary Material}

\author{First Author\\
Institution1\\
Institution1 address\\
{\tt\small firstauthor@i1.org}
\and
Second Author\\
Institution2\\
First line of institution2 address\\
{\tt\small secondauthor@i2.org}
}

\maketitle
\begin{abstract}
In this document, supplementing the main paper we present additional illustrations and visual results.
In addition, we also provide animations for the inference process on several images of the test set.
We find that these provide additional insights into what the model learns and where it is uncertain.
\end{abstract}

\begin{figure*}[h]
\includegraphics[trim=0 20mm 0 30mm,clip,width=1\linewidth]{inference_graph.pdf}
\caption{Inference visualization}
\end{figure*}
\vskip

\begin{figure*}[h]
\begin{subfigure}{1\linewidth}
    \centering
    \begin{subfigure}[t]{0.2\textwidth}    
  \includegraphics[width=\linewidth]{prior_visualization/horses/0_10.png}
    \end{subfigure}%
    ~
    \begin{subfigure}[t]{0.2\textwidth}    
  \includegraphics[width=\linewidth]{prior_visualization/horses/1_10.png}
    \end{subfigure}%
    ~
    \begin{subfigure}[t]{0.2\textwidth}    
  \includegraphics[width=\linewidth]{prior_visualization/horses/2_10.png} 
    \end{subfigure}%
    ~
    \begin{subfigure}[t]{0.2\textwidth}    
  \includegraphics[width=\linewidth]{prior_visualization/horses/3_10.png}
    \end{subfigure}%
    ~
    \begin{subfigure}[t]{0.2\textwidth}    
  \includegraphics[width=\linewidth]{prior_visualization/horses/4_10.png}
    \end{subfigure}%
\end{subfigure}
\begin{subfigure}[t]{1\linewidth}    
    \begin{subfigure}[t]{0.2\textwidth}    
  \includegraphics[width=\linewidth]{prior_visualization/horses/5_10.png}
    \end{subfigure}%
    ~
    \begin{subfigure}[t]{0.2\textwidth}    
  \includegraphics[width=\linewidth]{prior_visualization/horses/6_10.png}
    \end{subfigure}%
    ~
    \begin{subfigure}[t]{0.2\textwidth}    
  \includegraphics[width=\linewidth]{prior_visualization/horses/7_10.png}
    \end{subfigure}%
    ~
    \begin{subfigure}[t]{0.2\textwidth}    
  \includegraphics[width=\linewidth]{prior_visualization/horses/8_10.png}
    \end{subfigure}%
    ~
    \begin{subfigure}[t]{0.2\textwidth}    
  \includegraphics[width=\linewidth]{prior_visualization/horses/9_10.png}
    \end{subfigure}%
\end{subfigure}
\caption{Visualization of the learned shape priors. First 10 generates samples with $\sigma=10$. We observe that our model learns to incoorporate variability of the horse shapes, in particular in the legs.}
\label{fig:prior}
\end{figure*}